# Regularization-based Continual Learning for Anomaly Detection in Discrete Manufacturing

Benjamin Maschler [a,*], Thi Thu Huong Pham [a], Michael Weyrich [a]

[a] University of Stuttgart, Institute of Industrial Automation and Software Engineering, Pfaffenwaldring 47, 70569 Stuttgart, Germany

* Corresponding author. Tel.: +49 711 685 67295; Fax: +49 711 685 67302. E-mail address: benjamin.maschler@ias.uni-stuttgart.de

**Abstract**

The early and robust detection of anomalies occurring in discrete manufacturing processes allows operators to prevent harm, e.g. defects in production machinery or products. While current approaches for data-driven anomaly detection provide good results on the exact processes they were trained on, they often lack the ability to flexibly adapt to changes, e.g. in products. Continual learning promises such flexibility, allowing for an automatic adaption of previously learnt knowledge to new tasks. Therefore, this article discusses different continual learning approaches from the group of regularization strategies, which are implemented, evaluated and compared based on a real industrial metal forming dataset.

## 1. Introduction

Anomalies are a potential problem in all automated systems because they pose a challenge to any control software: By definition, they are what is not addressed by conventional rule or model based automation, and even in data-driven systems, handling them is difficult because they might occur unexpectedly and can differ considerably from any previous occurrence [1, 2]. However, simply ignoring anomalies can be detrimental with consequences ranging from inefficiencies [3] to complete failure [4], possibly harming workers or users [5].

A first step in anomaly handling is anomaly detection. Here, a shift from conventional, static methods towards deep learning based, dynamic approaches could be witnessed in recent years [1, 2]. However, even using those, data scarcity and high process dynamics remain challenging [6].

Mitigation could be provided by knowledge transfer between detection algorithms, bridging over gaps between different smaller datasets or various states of a process. Such a transfer could be realized using regularization-based continual learning approaches, which allow sequential learning of similar tasks without overwriting previously acquired knowledge [6].

*Objective*: In this article, the feasibility of different regularization approaches towards solving sequential learning problems is analyzed using a time series dataset taken from a discrete manufacturing process.

*Structure*: In chapter 2, related work on the topics of anomaly detection and regularization-based continual learning is presented. From there, a methodology is derived in chapter 3. Chapter 4 introduces a dataset and describes experiments conducted on that dataset as well as their results. Finally, chapter 5 draws a conclusion and givens an outlook.

## 2. Related work

### 2.1. Anomaly detection

Anomalies are a deviation from the rule or an irregularity that is not considered to be a part of the normal, intended system behavior. Anomalous dynamics are mostly unknown and occur inadvertently, lead to instabilities and are therefore drivers of increased inefficiencies and system errors [1, 2].

In literature, there are usually three different types of anomalies differentiated:

*Point anomalies* are characterized by a single data instance being considered anomalous. They are usually the easiest to detect and have been subject to extensive research [1, 2].

*Collective anomalies* are characterized by a series of data instances being considered anomalous as a group, although not each individual data instance needs to be anomalous. This necessarily requires relations between individual data instances, such as their sequence in time series data [1, 2].

*Contextual anomalies* are characterized by data being anomalous in specific, separately defined contexts only. This necessarily requires contextual attributes describing the data instances' context [1, 2].

Anomaly detection used to rely upon statistical, classification-based, clustering-based and information theoretic approaches. The common ground of those approaches is the fact that they aim to detect anomalies based on static, time-invariant models. However, many anomalies, especially collective and contextual ones, are dynamic and time-variant, calling for different approaches [2].

In recent years, deep learning based anomaly detection has started to fill this gap. Especially recurrent neural networks, such as long short-term memory (LSTM) and autoencoders (AE), have shown promising results:

In [5], a variational AE is used for anomaly detection based upon a comparison of log-likelihood real and reconstructed data. The algorithm outperforms state-of-the-art detectors on a dataset collected from a care robot tasked with feeding different people.

In [3], a stacked AE with interconnected gated recurrent units (GRU) for short-term and LSTM for long-term memory is used to detect anomalies as well as to predict the process outputs. The algorithm uniquely performs those different tasks simultaneously on a dataset collected from a multi-step hot forging process on hydraulic presses.

In [4], LSTM are combined with exponentially weighted moving average (EWMA) in order to significantly increase efficiency on contextual anomaly detection. This is demonstrated on a dataset collected from two simulated industrial robotic manipulators collaborating.

Although these examples highlight the suitability of deep learning based approaches towards solving real-life anomaly detection problems from the industrial domain, they still rely on large datasets in order to be trained. This limits their widespread, practical applicability.

## 2.2. Regularization-based continual learning

In the field of machine learning, the term 'continual learning' refers to the transfer of knowledge and skills from one or more source tasks to a target task in order to train a deep learning algorithm capable of solving both, source and target tasks [7, 8]. In the manufacturing domain, this can facilitate learning across several smaller, less homogenous datasets [6], mitigating two key problems hindering a more widespread utilization of machine learning [7]:

- Because of only small numbers of identical industrial machinery, high levels of data protection and little cooperation between different organizations, datasets sufficiently large and diverse for successful training are difficult to acquire [9].
- Because of a growing need for frequent reconfigurations [10], changing processes and dynamic environments, datasets once acquired only provide short-term representations of the problem space necessitating continuous data collection and retraining of algorithms [11].

Continual learning approaches are commonly divided into three categories: architectural, rehearsal and regularization approaches [12] (see Fig. 1). For the mitigation of the two problems described above, one of those is more promising than the other two: Whereas rehearsal approaches still rely on sharing of at least some data and architectural approaches strive only on more loosely related tasks, regularization approaches using altered loss functions in order to solve more closely related tasks appear best suited. *Regularization strategies* are modelled after synaptic consolidation in the brain, slowing down the change of certain weights depending on their importance on previously learned tasks, thereby selectively reducing the network's plasticity.

Four specific implementations of regularization strategies are commonly included in comparative analyses [12–14]:

*Elastic weight consolidation* (EWC) is based on the idea [15] that more than one set of weights $\theta$ represents a possible solution $\theta_A$ of a task A, so that a solution $\theta_{AB}$ can be found that solves both tasks A and B [16]. This is achieved by adding a penalty to the loss function (see Eq. 1): $L_C(\theta_{ABC})$ is the (conventional) loss for task C on a set of weights $\theta_{ABC}$ capable to solve all tasks A, B and C, $\lambda$ defines the importance of old tasks compared to the new one, $F$ is the diagonal of the Fisher information matrix and $i$ labels each individual parameter.

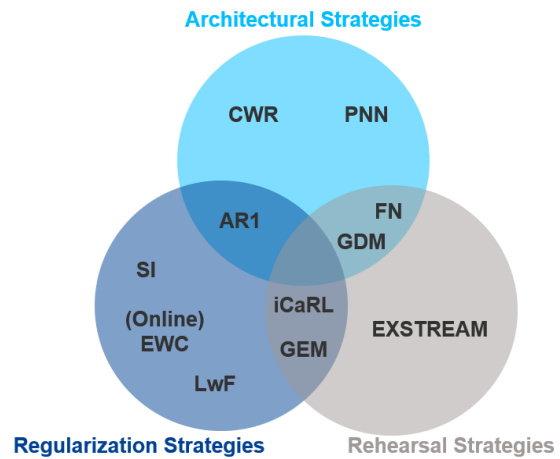

Fig. 1. Venn diagram of some of the most popular continual learning strategies based upon [12]

$$L(\theta_{ABC}) = L_C(\theta_{ABC}) + \lambda \cdot \sum_i [F_{A,i}(\theta_{ABC,i} - \theta_{A,i}^*)^2 + F_{B,i}(\theta_{ABC,i} - \theta_{B,i}^*)^2] \quad (1)$$

*Online EWC* expands on this idea, but shifts attention from older to newer tasks by not relying on Fisher information matrices for every tasks but on only one for all tasks combined [17]. This reduces Eq. 1 to Eq. 2:

$$L(\theta_{ABC}) = L_C(\theta_{ABC}) + \lambda \cdot \sum_i F_{AB,i}(\theta_{ABC,i} - \theta_{AB,i}^*)^2 \quad (2)$$

*Synaptic intelligence* (SI) relies on a similar idea, but uses an importance measure $\omega$ that is calculated directly during the stochastic gradient descent as opposed to the Fisher information matrix which needs to be calculated separately [18], leading to Eq. 3:

$$L(\theta_{ABC}) = L_C(\theta_{ABC}) + \lambda \cdot \sum_i \omega_{AB,i}(\theta_{ABC,i} - \theta_{AB,i}^*)^2 \quad (3)$$

*Learning without forgetting* (LwF) uses a slightly different approach: Differentiating between global weights $\theta_s$, weights specific to the old task(s) $\theta_o$ and those for the new task $\theta_n$, the algorithm tries to minimize both, the loss on the old task using $(\theta_s, \theta_o)$ and on the new task using $(\theta_s, \theta_n)$. The loss on the old task is brought into the training of the new task by using the Hessian matrix, a representation of individual weights importance for the overall result – similar to EWC's Fisher information matrix or SI's importance measure [19]. This yields Eq. 4:

$$\theta_s^*, \theta_o^*, \theta_n^* \leftarrow argmin\,(\lambda_o L_o(Y_o, \hat{Y}_o) + L_n(Y_n, \hat{Y}_n) + R(\hat{\theta}_s, \hat{\theta}_o, \hat{\theta}_n)) \quad (4)$$

Although the aforementioned regularization approaches displayed good results on different general evaluation datasets [13, 14], there are to the authors' knowledge no publications on their performance in the industrial domain – with one exception being EWC for fault prediction [6]. This article therefore aims to provide a first comparative analysis of those approaches on time series data taken from an actual manufacturing process.

## 3. Methodology

Based upon the good performance of recurrent neural networks described in chapter 2.1 combined with the desire to create a simple algorithm requiring neither vast computing resources nor extensive optimization, a multilayer LSTM-approach was chosen as base algorithm. This allows a focus on the different regularization methods and their respective performance. However, one must keep in mind that using a different, more sophisticated base algorithm might very well further increase the overall performance.

For this base algorithm, an input layer of dimension 3000 is connected to stacked LSTM-layers which lead to a two-node output layer (see Fig. 2).

Upon this base algorithm, the different regularization approaches are implemented. Because they only alter the respective loss functions, no changes to the base algorithm's network structure were necessary.

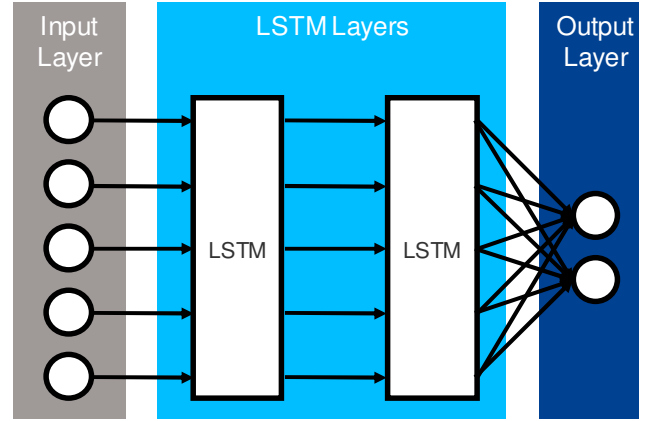

Fig. 2. Architecture of the base algorithm's deep neural network

Initial tests resulted in online EWC performing best. Therefore, a hyperparameter optimization for the base algorithm's parameters was carried out based upon this approach. Then, another hyperparameter optimization for the different approaches' parameters was conducted. The results are listed in Table 1.

In order to validate the different approaches' algorithm, tests were carried out on the permuted MNIST dataset. Our implementations achieved accuracies comparable to those reported in [14].

## 4. Experiments

In this chapter, a dataset collected from a discrete manufacturing process is introduced. The anomaly detection problem represented by the dataset can be classified as belonging to incremental domain learning [20]. Using this dataset, different experiments are conducted testing the previously described regularization approaches.

All experiments were conducted on a computer featuring an AMD Ryzen Threadripper 2920X CPU and a NVIDIA GeForce RTX 2080 8 GB GPU running Ubuntu 20.04. The learning framework used was PyTorch 1.6 under Python 3.6.

Table 1: Hyperparameters used for the different regularization-based continual learning algorithms

| Parameter | Value |
|---|---|
| Learning rate | 0.001 |
| Batch size | 100 |
| Number of hidden layers | 2 |
| Number of nodes per hidden layer | 200 |
| $\lambda_{EWC}$ | 1,000,000 |
| $\gamma_{EWC}$ | 10 |
| $c_{SI}$ | 300 |
| $T_{LwF}$ | 10 |
| $R_{LwF}$ | 1 |

*4.1. Experimental dataset*

The experiments were conducted using the subset of a very large industrial metal forming dataset collected on a hydraulic press. It consists of data from eight pumps applying pressure on a shared oil reservoir. Du to this setup, anomalous behavior of one pump is compensated by other pumps (see Fig. 3). This hides such behavior from the operator, because initially no problems occur. However, the other pumps experience increased wear, which makes an early detection of the described anomalous behavior desirable.

The challenge is further increased by frequent alterations of the production process, either by improvements such as new molds or changes of the manufactured product. Every alteration causes a change of the process' characteristics, which require anomaly detection algorithms to be retrained.

In the dataset used in this study, labeled pressure data from the production of fifteen different products each being produced hundreds of times are included.

*4.2. Regularization-based continual learning anomaly detection on a sequence of five tasks*

A first set of experiments was conducted randomly drawing twenty sequences of five tasks, i.e. five different products, and using all four approaches to classify normal or anomalous pump behavior. For baseline comparison, an algorithm without any regularization enhancement was used. After training on one task, the algorithm was validated on all tasks before training switched to the next task.

*Without regularization* (see Fig. 4, top left diagram), the accuracy on the task being trained on is about 0.92, whereas on the others it is between 0.5 and 0.7. The rapid decline of accuracy on one task after training switches to the next task, commonly called catastrophic forgetting [21], is easy to distinguish. Overall, even after training on all tasks, the algorithm is clearly not capable of solving all of them sufficiently well, underlining the need for a different approach.

When *EWC* is used (see Fig. 4, top middle diagram), the accuracy on the task being trained on is slightly declining from 0.92 for task 1 to 0.9 for task 5. Prior to being trained on, a task's accuracy lies between 0.5 and 0.65. After training switches from on task to the next, the prior tasks' accuracies decline – although considerably less than without any regularization.

When *Online EWC* is used (see Fig. 4, top right diagram), the accuracy on the task being trained on is considerably declining from 0.92 for task 1 to about 0.8 for task 5. Prior to being trained on, a task's accuracy lies between 0.5 and 0.65. After training switches from on task to the next, the prior task's accuracy declines, but stays somewhat constant following subsequent switches. This causes the final accuracies of all tasks except the last being trained on to be significantly higher than without any regularization or using EWC.

When *LwF* is used (see Fig. 4, bottom right diagram), the resulting accuracies resemble that without any regularization with an average offset of about +0.025. This offset is caused by slight accuracy improvements on previously trained on tasks, while the accuracies on the tasks currently being trained are similar.

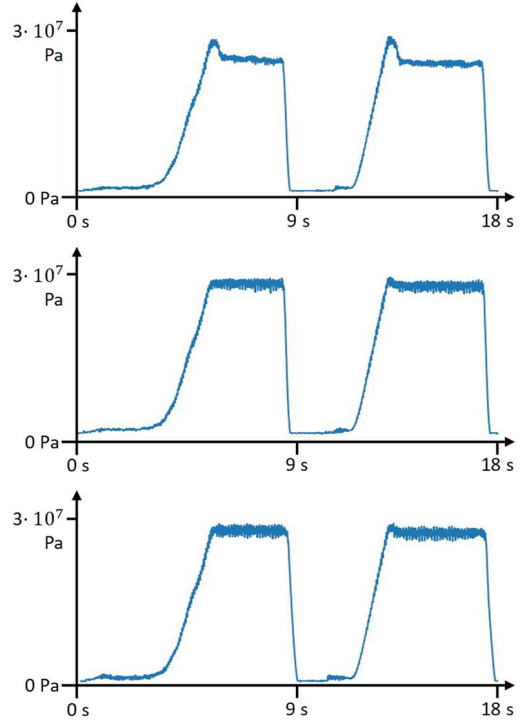

Fig. 3. Comparison of data from normal (middle and bottom) and anomalous (top) pumps

When *SI* is used (see Fig. 4, bottom middle diagram), the resulting accuracies also resemble that without any regularization with an increasing average offset of -0.01 for tasks 1 to +0.025 for task 5. This time, the offset is not caused by retained skills.

Comparing the mean accuracies of the different approaches (see Fig. 4, bottom left diagram), no regularization, LwF and SI result in similar curves, distinguishable mainly due to the previously described offsets. Starting at about 0.62 they peak during training on tasks 3 and 4 before slightly declining again. Contrastingly, EWC's and Online EWC's accuracies keep on rising throughout the entire training sequence.

Table 2 focusses on the different tasks' accuracies during the training of task 5: Although it has the overall highest accuracy on task 5, no regularization expectably fares worst regarding the lowest (0.52 for task 2) and the mean accuracies (0.66). LwF and SI perform only marginally better. With only a slightly worse accuracy on task 5 (0.9), EWC has much better

Table 2: Final accuracy on five tasks after training on five tasks

| Approach | Best | Mean | Worst |
|---|---|---|---|
| No Regularization | **0.93** | 0.66 | 0.52 |
| Elastic Weight Consolidation | 0.9 | 0.77 | 0.67 |
| Online Elastic Weight Consolidation | 0.85 | **0.82** | **0.79** |
| Learning without Forgetting | 0.92 | 0.68 | 0.57 |
| Synaptic Intelligence | 0.91 | 0.68 | 0.57 |

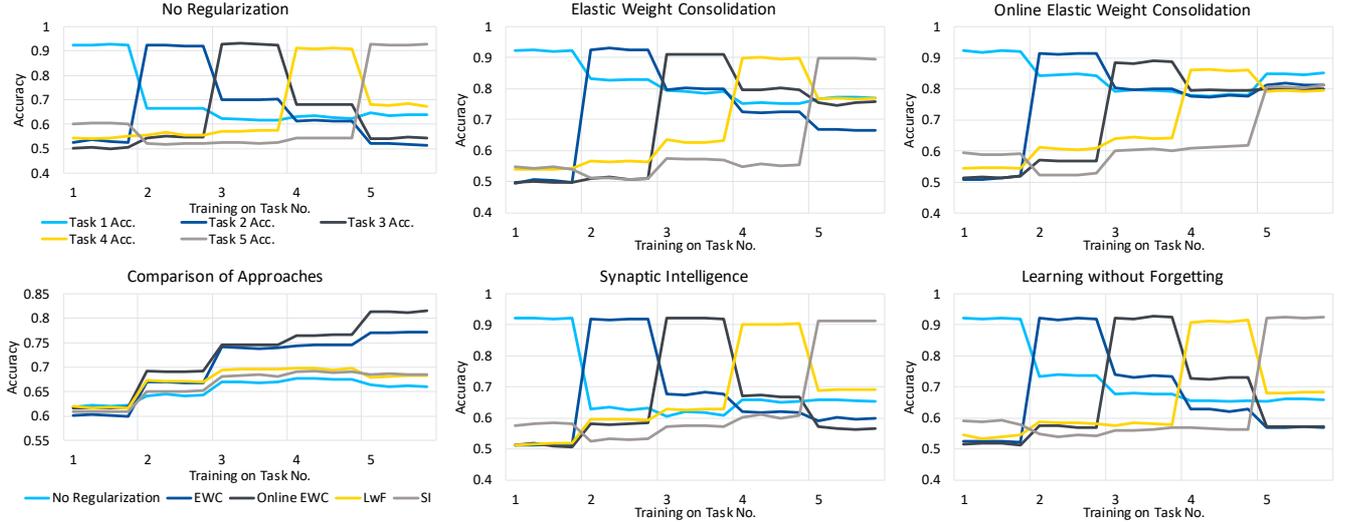

Fig. 4. Results on a transfer learning scenario involving twenty randomly drawn sequences of five different products using no regularization, EWC, online EWC, LwF or SI approaches (clockwise, starting top left; legend only in first diagram) as well as a comparison of mean accuracies (bottom left)

lowest (0.67 for task 2) and mean (0.77) accuracies. Concludingly, Online EWC combines the worst accuracy on task 5 (0.85) with best worst (0.79 for task 4) and mean (0.82) accuracies.

Online EWC clearly performs best in learning to solve all five tasks. After having trained on all of them, all tasks' accuracies are in the vicinity of 0.8. However, the decreasing accuracy on the task being currently trained on raises the question of its performance when more tasks are added to the sequence.

*4.3. Regularization-based continual learning anomaly detection on a sequence of eight tasks*

A second set of experiments repeats the first set using eight tasks being trained on sequentially instead of five. The key characteristics remain similar:

*Without regularization* (see Fig. 5, top left diagram), the accuracy on the task being trained on is about 0.9 to 0.92, whereas on the others it is between 0.4 and 0.75 with distinct catastrophic forgetting. Again, *LwF* (see Fig. 5, bottom right diagram) and *SI* (see Fig. 5, bottom middle diagram) perform slightly better with the notable exceptions of SI's performance on tasks 4 (+0.05) and 7 (+0.065).

When *EWC* is used (see Fig. 5, top middle diagram), the accuracy on the task being trained on lies between 0.85 and 0.91 without a clear trend. The effect of catastrophic forgetting is reduced significantly.

Table 3: Final accuracy on eight tasks after training on eight tasks

| Approach | Best | Mean | Worst |
|---|---|---|---|
| No Regularization | **0.93** | 0.61 | 0.50 |
| Elastic Weight Consolidation | 0.88 | 0.73 | 0.64 |
| Online Elastic Weight Consolidation | 0.82 | **0.75** | **0.68** |
| Learning without Forgetting | 0.93 | 0.62 | 0.50 |
| Synaptic Intelligence | 0.88 | 0.62 | 0.54 |

When *Online EWC* is used (see Fig. 5, top right diagram), the accuracy on the task being trained on is considerably declining from 0.9 to 0.68. Again, accuracy levels stabilize after initial drops once the respective task is no longer being trained on. However, for the last task, training does not even bring it to the accuracy level of the other tasks after their drop. Therefore, the assumption that the regularization capacity of our implementation is depleted after eight tasks appears to be correct.

The significant differences in mean accuracies between the first five training phases of the second set of experiments (see Fig. 5, bottom left diagram) and the first set of experiments hints towards a strong influence of individual tasks on the overall results despite averaging over twenty randomly created task sequences.

Table 3 gives the different tasks' accuracies during training of task 8. Compared with Table 2, the absolute accuracies decreased slightly, but the relative order remained without significant changes, leaving online EWC performing best over all eight tasks.

## 5. Conclusion

In this paper, the feasibility of different regularization approaches towards solving sequential learning problems in industrial use cases was examined. A time series dataset taken from a discrete manufacturing process was used to evaluate the algorithms.

Our main findings are:
- Regularization approaches improve our base algorithm's performance compared to no regularization.
- Online elastic weight consolidation outperforms elastic weight consolidation, learning without forgetting and synaptic intelligence.
- The performance of all approaches decreases with the number of tasks to be learnt.

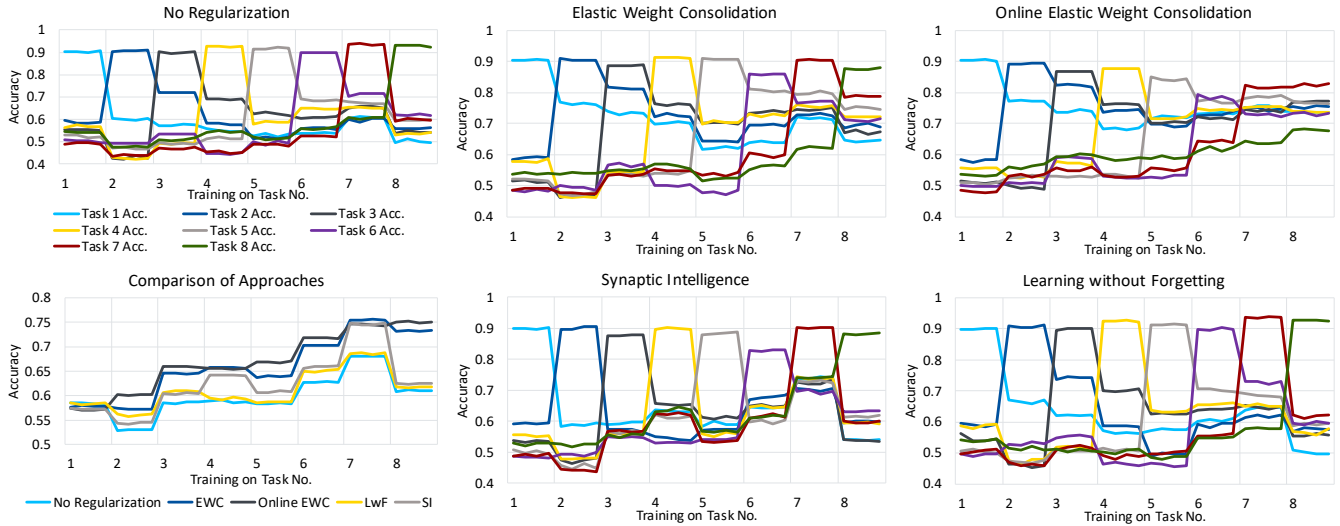

Fig. 5. Results on a transfer learning scenario involving twenty randomly drawn sequences of eight different products using no regularization, EWC, online EWC, LwF or SI approaches (clockwise, starting top left; legend only in first diagram) as well as a comparison of mean accuracies (bottom left)

- The tasks themselves have a big influence on the learning performance. It remains unclear, whether this only depends on their relative similarity to each other or also on their individual position in the learning sequence.

Future research should carry out hyperparameter optimizations for all approaches, possibly even for different task sequence lengths. Furthermore, the number of randomly created task sequences should be increased in order to reduce the impact of each individual sequence. Additionally, experiments involving other industrial datasets could be of interest.

**Acknowledgment**

This work was supported by the Otto Fuchs KG who provided the data.

**References**

[1] Chandola V, Banerjee A, Kumar V. Anomaly detection. ACM Comput. Surv. 2009; 3:1–58.
[2] Lindemann B et al. A Survey on Anomaly Detection with LSTM Networks. Preprint.
[3] Lindemann B, Jazdi N, Weyrich M. Anomaly detection and prediction in discrete manufacturing based on cooperative LSTM networks. 2020 IEEE 16th International Conference on Automation Science and Engineering (CASE), Hong Kong; 2020; 1003–10.
[4] Ding S et al. Model-Based Error Detection for Industrial Automation Systems Using LSTM Networks. 7th International Symposium on Model-based Safety and Assessment (IMBSA 2020), Lisbon; 2020; 212–26.
[5] Park D, Hoshi Y, Kemp C. A Multimodal Anomaly Detector for Robot-Assisted Feeding Using an LSTM-Based Variational Autoencoder. IEEE Robot. Autom. Lett. 2018; 3:1544–51.
[6] Maschler B, Vietz H, Jazdi N, Weyrich M. Continual Learning of Fault Prediction for Turbofan Engines using Deep Learning with Elastic Weight Consolidation. 2020 25th IEEE International Conference on Emerging Technologies and Factory Automation (ETFA), Vienna; 2020; 959–66.
[7] Maschler B, Weyrich M. Deep Transfer Learning for Industrial Automation: A Review and Discussion of New Techniques for Data-Driven Machine Learning. Industrial Electronics Magazine 2021; (accepted).
[8] Parisi G et al. Continual lifelong learning with neural networks: A review. Neural Networks 2019; 113:54–71.
[9] Assadi A et al. Survey on Sharing Data for Machine Learning in Industry. Preprint: 2020.
[10] Müller T, Schmidt J-P, Jazdi N, Weyrich M. Cyber-Physical Production Systems: Enhancement with a Self-Organized Reconfiguration Management. 2020 14th CIRP Conference on Intelligent Computation in Manufacturing Engineering (ICME), Gulf of Naples; 2020;
[11] Tercan H, Guajardo A, Meisen T. Industrial Transfer Learning: Boosting Machine Learning in Production. Proceedings of the 2019 IEEE 17th International Conference on Industrial Informatics (INDIN), Helsinki; 2019; 274–9.
[12] Maltoni D, Lomonaco V. Continuous learning in single-incremental-task scenarios. Neural Networks 2019; 116:56–73.
[13] Hsu Y-C, Liu Y-C, Ramasamy A, Kira Z. Re-evaluating Continual Learning Scenarios: A Categorization and Case for Strong Baselines. 32nd Conference on Neural Information Processing Systems (NeurIPS) Continual Learning Workshop 2018.
[14] van de Ven G, Tolias A. Three scenarios for continual learning. 32nd Conference on Neural Information Processing Systems (NeurIPS) Continual Learning Workshop 2018.
[15] Hinton G, Vinyals O, Dean J. Distilling the Knowledge in a Neural Network. 28th Conference on Neural Information Processing Systems (NeurIPS) Continual Learning Workshop 2014.
[16] Kirkpatrick J et al. Overcoming catastrophic forgetting in neural networks. Proceedings of the National Academy of Sciences of the United States of America 2017; 13:3521–6.
[17] Schwarz J et al. Progress & Compress: A scalable framework for continual learning. Proceedings of Machine Learning Research 2018; 80:4528–37.
[18] Zenke F, Poole B, Ganguli S. Continual Learning Through Synaptic Intelligence. Proceedings of Machine Learning Research 2017; 70:3987–95.
[19] Li Z, Hoiem D. Learning without Forgetting. IEEE transactions on pattern analysis and machine intelligence 2018; 12:2935–47.
[20] Maschler B, Kamm S, Weyrich M. Deep Industrial Transfer Learning at Runtime for Image Recognition. at - Automatisierungstechnik 2021; (accepted).
[21] French R. Catastrophic forgetting in connectionist networks. Trends in Cognitive Sciences 1999; 4:128–35.